\documentclass[11pt,a4paper]{article}
\usepackage[hyperref]{emnlp2020}
\usepackage[utf8]{inputenc}
\usepackage{times}
\usepackage{latexsym}
\usepackage{amsmath}
\usepackage{amssymb}
\usepackage{graphicx}
\usepackage{xcolor}
\usepackage{xifthen}
\usepackage{multirow}
\usepackage{xspace} % add or not a space
\usepackage{subfig}
\usepackage{colortbl}
\usepackage{arydshln} %dashed vertical lines
\usepackage{tipa}
\usepackage{linguex}
\usepackage{textcomp}

\usepackage{diagbox}
\usepackage{float}
\usepackage{multicol}

\newcommand{\plh}{%
  {\ooalign{$\phantom{0}$\cr\hidewidth$\scriptstyle\times$\cr}}%
}

\usepackage{newunicodechar}
\DeclareFontEncoding{LS1}{}{}
\DeclareFontSubstitution{LS1}{stix}{m}{n}
\DeclareFontFamily{LS1}{stixscr}{\skewchar\font127 }
\DeclareFontShape{LS1}{stixscr}{m}{n} {<->s*[.7] stix-mathscr}{}
\newunicodechar{◌}{{\usefont{LS1}{stixscr}{m}{n}\symbol{\string"E3}}}

\newcommand{\all}{\textsc{all}\xspace}
\newcommand{\keep}{\textsc{keep}\xspace}
\newcommand{\true}{\textsc{true}\xspace}
\newcommand{\random}{\textsc{random}\xspace}
\newcommand{\baseline}{\textsc{baseline}\xspace}

\newcommand{\grupackager}[1][]{%
    \ifthenelse{\isempty{#1}}
    {$\textsc{GRU}{_\textsc{PACK.}}$\xspace}
    {{$\textsc{GRU}{_\textsc{PACK.}}\text{--#1}$}\xspace}%
}

\usepackage{microtype}
\aclfinalcopy % Uncomment this line for the final submission
\def\aclpaperid{101} %  Enter the acl Paper ID here

\title{Catplayinginthesnow: Impact of Prior Segmentation on a Model of Visually Grounded Speech}

\author{
   {William N. Havard$^{1,2}$,
   Laurent Besacier$^{1}$,
   Jean-Pierre Chevrot$^{2}$}\\\\
   {$^{1}$ LIG, Univ. Grenoble Alpes, CNRS, Grenoble INP, 38000 Grenoble, France}\\
   {$^{2}$ LIDILEM, Univ. Grenoble Alpes, 38000 Grenoble, France}\\
   {\tt\small first-name.lastname@univ-grenoble-alpes.fr}
}

\date{\today}

\begin{document}

% Linguex margin modification
\setlength{\SubExleftmargin}{1em}
\setlength{\Exlabelsep}{0.5pt}
\setlength{\Exindent}{1em}
% End linguex margin modification

\maketitle

\begin{abstract}
The language acquisition literature shows that children do not build their lexicon by segmenting the spoken input into phonemes and then building up words from them, but rather adopt a top-down approach and start by segmenting word-like units and then break them down into smaller units. This suggests that the ideal way of learning a language is by starting from full semantic units. In this paper, we investigate if this is also the case for a neural model of Visually Grounded Speech trained on a speech-image retrieval task. We evaluated how well such a network is able to learn a reliable speech-to-image mapping when provided with phone, syllable, or word boundary information. We present a simple way to introduce such information into an RNN-based model and investigate which type of boundary is the most efficient. We also explore at which level of the network's architecture such information should be introduced so as to maximise its performances. Finally, we show that using multiple boundary types at once in a hierarchical structure, by which low-level segments are used to recompose high-level segments, is beneficial and yields better results than using low-level or high-level segments in isolation.
\end{abstract}

\section{Introduction and Prior Work}

Visually Grounded Speech (VGS) models whether CNN-based \cite{Flickr8k_Audio, harwath_2016, kamper_vgs} or RNN-based \cite{Chrupala2017, Merkx2019LanguageLU} became recently popular as they enable to model complex interaction between two modalities, namely speech and vision, and can thus be used to model child language acquisition, and more specifically lexical acquisition. Indeed, these models are trained to solve a speech-image retrieval task. That is, given a spoken input description, they are trained to retrieve the image that matches the description the best. This task requires the model to identify lexical units that might be relevant in the spoken input, detect which objects are present in the image, and finally see if those objects match the detected spoken lexical units. Their task is thus very close to that of a child learning its mother tongue, who is surrounded by a visually perceptible context and tries to match parts of the acoustic input to surrounding visible situations. Research in language acquisition have put forward that children do not build their lexicon by segmenting the spoken input into phonemes and then building up words, but rather adopt a top-down approach \cite{bortfeld_top_down} and start by identifying and memorising whole words \cite{Jusczyk1995} or chunks of words \cite{bannard_matthews_chunks} and then segment the spoken input into smaller units, such as phonemes. This suggests that the most efficient way of segmenting the spoken input to map a visual context to its description is at word level.
From a more technological point of view, speech-based models lag behind their textual counterparts. For example, speech-image retrieval performs worse than text-image retrieval, despite being trained on the same data, the only changing factor being the modality where text or speech is used as a query. 
This begs the question: what makes text inherently better than speech for such applications? Is it because text is made up of already-segmented (discrete) units which lack internal variation, or because these discrete units (usually tokens) stand for full semantic units, or a combination of both?

Since the pioneering computational modelling work of lexical acquisition by \citet{Roy02_CELL}, neural network enabled  an even tighter interaction between the visual and the audio modalities.
Recent works suggest that  networks trained on a speech-image retrieval task perform an implicit segmentation of their input. Whether CNN-based approaches or RNN-based approaches are employed, all seem to segment individual words from the inputted spoken utterance \cite{harwath_2016, Chrupala2017, Havard2019, Havard_CoNLL_2019, Merkx2019LanguageLU}. This result stands also for languages other than English, such as Hindi or Japanese \cite{harwath_interlingua, Havard2019, Azuh2019, harwath_japanese}.
\citet{Chrupala2017} and \citet{Merkx2019LanguageLU}, however, observed  that not all layers encode word-like units, suggesting that some layers specialise in lexical processing whereas some other do not encode such information.

\textbf{Contributions} Our research question can be framed as follows: what is the segmentation that maximises the performance of an audio-visual network if speech were to be segmented? To answer this question we investigate \emph{how} it is possible to give speech boundary information to a neural network and explore \emph{which} type of boundary (phone, syllable, or word) is the most efficient. We also explore \emph{where} such information should be provided, that is, at which layer of the architecture is the addition of this information the most beneficial?

\section{Model \& Data}
\label{sec:model-data}
\textbf{Data }
We use two different data sets in our experiments: MS~COCO \cite{MSCOCO} and Flickr8k \cite{Flickr8k_Images}. Both corpora were initially conceived for computer vision purposes and both feature a set of images along with five written descriptions 
of the images. The captions were not computer generated but written by humans. We use the audio extensions of both data sets: for Flickr8k, we use the captions provided by \citet{Flickr8k_Audio}, and for COCO we use Synthetic COCO data set introduced by \cite{Chrupala2017, SynthCOCO}. The captions of \citet{Flickr8k_Audio} were gathered using Amazon Mechanical Turk and were thus uttered by humans. This data set is particularly challenging as it features multiple speakers and the quality of the recording is uneven from one caption to another. The spoken captions of \citet{SynthCOCO} feature synthetic speech generated with Google's Text-to-Speech system. 
For both corpora, we extracted speech-to-text alignments through the \textit{Maus} forced aligner
~\cite{WebMaus} online platform, resulting in alignments at word and phone levels.

\textbf{Architecture } The models we train in
our experiments all have the same architecture and are based on that of \citet{Chrupala2017}.\footnote{The code we use is based on \url{https://github.com/gchrupala/vgs}} As all models of VGS, be they CNN-based or RNN-based, this architecture has two main components: an image encoder and a speech encoder. Such models are trained to solve a speech-image retrieval task, that is, given 
a query in the form of a spoken description, they should retrieve the closest matching image fitting the description.

The image encoder is a simple linear layer that reduces pre-computed VGG image vectors to the desired dimension. The speech encoder, which receives MFCC vectors as input, consists of a 1D convolutional layer, followed by five stacked recurrent layers with residual connections, followed by an attention mechanism. We use uni-directional recurrent layers and not bi-directional recurrent layers even though it has been shown they lead to better results \cite{Merkx2019LanguageLU}. Indeed, we aim at having a cognitively plausible model: humans process speech in a left-to-right fashion, as speech is being gradually uttered, and not from both ends simultaneously.
We use the same loss function as initially used by \citet{Chrupala2017}:
\begin{equation}
  \resizebox{1\hsize}{!}{
  \begin{math}
    \begin{split}
    \mathcal{L}(u, i, \alpha)=
      \sum_{u, i} 
      \Bigg( \sum_{u'}\max [0, \alpha + d(u, i) - d(u', i)]\\
      +\sum_{i'}\max[0, \alpha + d(u, i) - d(u, i')] \Bigg)
    \end{split}
  \end{math}
  }
  \label{eq:loss_fct}
\end{equation}
This contrastive loss function encourages the network to minimise the cosine distance $d$ by a margin $\alpha$  between an image $i$ and its corresponding utterance $u$, while maximising the distance between mismatching image/utterance pairs $i'$/$u$ and $i$/$u'$. In our experiments we set $\alpha=0.2$.
 \begin{figure*}[h!]
	\begin{minipage}{.331\linewidth}
		\centering
		\subfloat[]{\label{fig:VanillaGRU}\includegraphics[width=0.98\textwidth,height=0.8\textwidth]{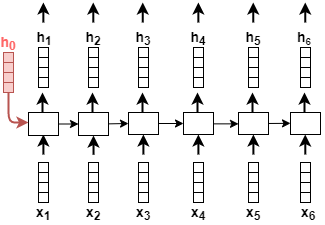}}
	\end{minipage}
	\vline
	\begin{minipage}{.332\linewidth}
		\centering
		\subfloat[]{\label{fig:AllGRU}\includegraphics[width=0.98\textwidth,height=0.8\textwidth]{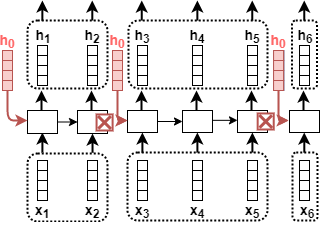}}
	\end{minipage}
	\vline
	\begin{minipage}{.332\linewidth}
		\centering
		\subfloat[]{\label{fig:KeepGRU}\includegraphics[width=0.98\textwidth,height=0.8\textwidth]{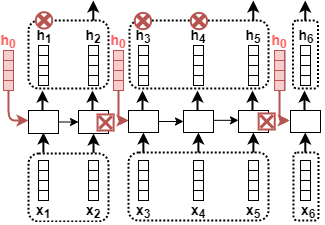}}
	\end{minipage}
	\caption{Graphical representation of the different GRUs used in our experiments. Figure~\ref{fig:VanillaGRU} shows a Vanilla GRU. Figure~\ref{fig:AllGRU} shows \grupackager in the \all condition where all the vectors produced at each time step are passed on to the next layer. \ref{fig:KeepGRU} shows \grupackager in the \keep condition where only the last vector of a segment is passed on to the next layer, thus resulting in a output sequence shorter than the input sequence. The red crosses inscribed in a square (\raisebox{-.2\height}{\includegraphics[scale=0.6]{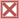}}) signal that the output vector computed at a given timestep is not passed on to the next timestep and that the initial state $h_0$ is passed on instead. The red crosses inscribed in a circle (\raisebox{-.2\height}{\includegraphics[scale=0.6]{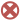}}) signal that the output vector computed at a given timestep is not passed on to the next layer. Dotted line group vectors belonging to a same segment (either phone, syllable-connected, syllable-word, or word). Note that $h_0$ is only passed on to the next state at the end of a segment, thus effectively materialising a boundary by manually resetting the history. Also note that the $x_1, x_2, ..., x_{t}$ figured in this representation could either be the original input sequence (in our case, acoustic vectors) or could also be the output of the previous recurrent layer.}
	\label{fig:gru}
\end{figure*}

\textbf{Hyperparameters} For both COCO and Flickr8k we use 1D convolutions with 64 filters of length 6 and a stride of 1 to preserve the original time resolution (and hence, boundary position).
We use 512 units per recurrent layer for COCO and 1024 for Flickr8k. All models were trained using Adam optimiser and an initial learning rate of $0.0002$. For our experiments we use the pre-computed MFCC vectors and pre-computed VGG vectors provided by \citet{Chrupala2017}.\footnote{12 MFCC coef. plus energy for COCO; 12 MFCC coef. plus energy as well as deltas and delta deltas for Flickr8k.} We also use the same training, validation and testing splits.\footnote{Training/Validation/Test split contain 113,287/5,000/5,000 images (COCO) and 6,000/1,000/1,000 images (Flickr8k).} 

\section{Integrating Segmentation Information}
\label{sec:integrate-boundary-information}
\subsection{Boundary Types}
\label{sub-sec:boundary-types}
As previously stated, we are interested in supplying our network with linguistic information such as segment boundaries. We define a segment as either being a phoneme, a syllable, or a word. 
We consider two different types of syllables. Indeed, when we speak, words are not uttered one after the other in a disconnected fashion, but are rather blended together through a process called ``resyllabification''. In English, this phenomenon is visible when a word ending with a consonant is followed by a word starting with a vowel. In this case, the final consonant of the first word tends to be detached from it and attached to the next word, thus crossing the word boundary. This phenomenon is illustrated in Example~\ref{ex:resyllabification} where phonemes in red indicate a resyllabification phenomenon.

\ex.\label{ex:resyllabification}
    \hfill{This~is~an~article.}\\
    \textit{Transcription}\footnote{We use ``\#'' to signal word boundaries and ``.'' to signal syllable boundaries.}\hfill {/\textipa{DIs}\#\textipa{Iz}\#\textipa{@n}\#\textipa{A\*rtIk@l}/}
    \a. \label{ex:wo-resyllabification}\textit{No resyllabification}\hfill {/\textipa{DIs.Iz.@n.A\*r.tI.k@l}/}
    \b. \label{ex:w-resyllabification} \textit{With resyllabification}\hfill {/\textipa{DI.\textcolor{red}{s}I.\textcolor{red}{z}@.\textcolor{red}{n}A\*r.tI.k@l}/}
\par

For the rest of this article ``syllables-word" will refer to syllables that result of a segmentation that does not take into account resyllabification \ref{ex:wo-resyllabification}, whereas ``syllables-connected'' will refer to syllables that result of a segmentation that takes into account resyllabification \ref{ex:w-resyllabification}. It should be noted that in the syllables-connected condition, most word boundaries are lost.\footnote{Word boundaries are not lost in the following cases: V\#V and C\#C when CC is not an allowed complex onset. C and V respectively refer to ``consonant" and ``vowel".} In the syllables-word condition, however, all word boundaries are preserved and the segmentation inside a word may 
result in a morphemic segmentation (as for example in ``runway" /\textipa{\*r2n.weI}/ or ``air.plane" /\textipa{E\*r.plein}/). Nevertheless, this is not always the case, especially for longer words that are of non-germanic origin (such as ``elephant" /\textipa{E.lE.fant}/ or ``computer" /\textipa{k@m.pju.t\textschwa\textrhoticity}/). We expect models trained in the syllables-connected condition to perform worse than those trained in the syllables-word condition as resyllabification hinders word recognition \cite{Vroomen1999}.

Segment boundaries were derived from the forced alignment metadata 
so as to indicate which MFCC vector constitutes a boundary or not.\footnote{As the force aligner used does not provide alignment at the syllable level, we wrote a custom script to recreate syllables from the phonemic transcription.} Therefore, for each caption we have a sequence $X$ of length $T$ of $d$-dimensional acoustic vectors $X=~\left[x_{1}^d, x_{2}^d, ..., x_{T}^d\right]$ and a corresponding sequence of scalars $B$ 
representing boundaries $B=~\left[b_{1}, b_{2}, ..., b_{T}\right]$, $b_{t}\in\left\{0,1\right\}$, where $b_{t}\triangleq 1$ if $x_{t}$ is a segment boundary, $0$ otherwise.

\subsection{Integrating Boundary Information}
\label{sec:integrate}
In order to integrate boundary information into the model, we take advantage of how recurrent neural networks compute their output. 
They can be formalised as follows:
\begin{equation}
h_{t} = f\left(h_{t-1}, x_{t}; \theta\right)
\end{equation}
where the hidden state at timestep $t$ noted $h_t$ is a function $f$ of the previous hidden state $h_{t-1}$ and the current input at $x_{t}$, with $\theta$ being learnable parameters of the function $f$. A special case arises at the very first time step $t=1$ as $h_{t-1}$ does not exist. In this case, the initial state $h_{t-1}$ noted $h_0$ is set to be a vector of $0$.  The output of such a network at timestep $T$ is thus dependent on all the previous timesteps. An illustration of such a network is depicted in Figure~\ref{fig:VanillaGRU}. In this work, we use GRUs \cite{GRU}, but our methodology is applicable to any other type of recurrent cell such as vanilla RNNs or LSTMs.

Our approach to integrate boundary information into the network can be formalised as follows:
\begin{align}
 h_t =\begin{cases} 
            f\left(h_{0}, x_{t}; \theta\right), \text { if }b_{t-1}=1\\
            f\left(h_{t-1}, x_{t}; \theta\right), \text{ otherwise}
        \end{cases}\label{eq:halting_prob}
\end{align}
In our approach, $h_t$ is only dependent on the previous timestep $h_{t-1}$ if the previous timestep was not an acoustic vector corresponding to segment boundary ($b_{t-1}\ne~0$). If the previous timestep corresponds to a segment boundary ($b_{t-1}=1$), we reset the hidden state so that it is equal to $h_0$. Hence, vectors in the same segment are temporally dependent, but vectors belonging to two different segments are not. The GRUs that use this computing scheme will from now on be referred to as \grupackager, as vectors belonging to the same segment are ``packed" together.

We derived two different conditions from this initial setting: \all and \keep. In the \all condition (see Figure~\ref{fig:AllGRU}), all the vectors belonging to a segment are forwarded to the next layer (which can either be a recurrent layer, or an attention mechanism depending on the position of the \grupackager layer.) In the \keep condition, only the last vector of each segment is forwarded to the next layer (see Figure~\ref{fig:KeepGRU}). The length of the output and input sequence stays the same in the \all condition. However, it should be noted that in the \keep condition, the length of the output sequence is shorter than the input sequence. Potentially, the length of the sequences can be different for different items inside a batch as the captions have a different number of segments (be they phones, syllables or words). For this reason, and as the subsequent layers expect a 3D rectangular matrix,\footnote{Of shape $\text{batch size}\times\text{sequence}\times\text{embedding dimension}$.} we add padding vectors on the sequence dimension until all the elements of the batch have the same sequence length.
The difference between \all and \keep is motivated by the fact that we believe that keeping the last vector of a segment could constrain the network to build more consistent representations for different occurrences of the same segment, as the subsequent layers will have less information to rely on.
A similar approach to ours was proposed by \citet{audio_word2vec} in an Audio-Word2Vec experiment, where instead of being given gold segment boundaries, a classifier outputs a probability that a given frame constitutes a segment boundary.

\section{Experiments and Results}
\label{sec:exp-settings}

\subsection{\grupackager Position and Random Boundaries}
\label{sub-sec:grupack-position}
In order to understand where boundary information should be introduced (that is, at which level of the architecture), we train as many models as the number of recurrent layers, where each time one layer of GRUs is replaced with one \grupackager layer. For example, ``\grupackager[3]'' refers to a model where the third layer of GRUs is a \grupackager layer and other layers ($1^{st}$,$2^{nd}$,$4^{th}$,~and $5^{th}$ layer) are vanilla GRU layers. This setting will allow to explore \emph{where} introducing boundary information is the most efficient.

To understand if introducing boundary information helps the network in its task, we compare the performance of the models using boundary information with a baseline model which does not use any (thus, all the recurrent layers of the baseline architecture are Vanilla GRU layers). This model will from now on be referred to as \textsc{baseline}. We also introduce another condition, where, instead of training models with real segment boundaries (which from now on will be referred to as \textsc{true}), we train models with random boundaries (which from now on will be referred to as \textsc{random}). Indeed, it could be that randomly slicing speech into sub-units leads to better results, even though the resulting units do not constitute linguistically meaningful units. Consequently, training models with random boundaries will enable us to verify this claim. Random boundaries were generated by simply shuffling the position of the real boundaries (vector $B$ introduced in §\ref{sub-sec:boundary-types}), resulting in as many randomly positioned boundaries as there are real ones. Note that we do still expect the models to have reasonable results even when using random boundaries, as acoustic vectors are kept untouched. Nonetheless, we expect that placing random boundaries will hinder network's learning process and thus yield results significantly lower than when using true boundaries. We expect results to be significantly lower in the \random-\keep condition as this condition is equivalent to randomly subsampling the input, and thus removing a lot of information.

\begin{table}[tbp]
  \centering
  %\resizebox{0.40\textwidth}{!}{
    \begin{tabular}{c|c|c|c}%|c}
      \textbf{Data set} & \textbf{R@1} & \textbf{R@5} & \textbf{R@10} \\\hline%& \textbf{$\widetilde{r}$}
      COCO     & 9.0 & 27.0 & 39.5 \\\hline%& 17
      Flickr8k & 4.3 & 13.4 & 21.4 \\%& 54
    \end{tabular}
  %}
  \caption{
    Mean recalls at 1, 5, and 10 (in \%) on a speech-image retrieval task COCO and Flickr8k in the \baseline condition. Chance scores are 0.0002/0.001/0.002 for COCO and 0.001/0.005/0.01 for Flickr8k.
    }
  \label{tab:baseline-results}
\end{table}
\subsection{Evaluation}
\label{subsec:evaluation}
Models are evaluated in term of Recall@k (R@k). Given a spoken query, R@k evaluates the models ability to rank the target paired image in the top $k$ images. In order to evaluate if the results observed in our different experimental conditions (\true-\all, \true-\keep, \random-\all, \random-\keep) are different from one another and from the \baseline condition, we used a two-sided proportion Z-Test. This test is used to check if there is a statistical difference between two independent proportions. As for each spoken query there is only one target image, R@k becomes a binary value which equals $1$ if the target image is ranked in the top $k$ images and $0$ otherwise. In our case, the proportion that we test is the number of successes over the number of trials (which corresponds to the number of different caption/image pairs in the test set).

\subsection{Results}
Overall, our experimental settings led to the training of 81 different models per data set.\footnote{$(\text{Seg. type}~\in~\text{\{phone,syl.-connected,syl.-word,word\}}\allowbreak{\plh}~\text{\grupackager}\text{\{1,2,3,4,5\}}~{\plh}~\text{\{\textsc{true},\textsc{random}\}}~{\plh}~\text{\{\textsc{all},\textsc{keep}\}})\allowbreak+\textsc{baseline}$}
\baseline results are shown in Table~\ref{tab:baseline-results}, results for the \true/\random conditions obtained on the Flickr8k are shown in Table~\ref{tab:flickr-results} ~and results on COCO in Table~\ref{tab:coco-results} (Appendix \ref{sec:appendix-coco}). We obtain lower results on Flickr8k than on COCO which shows how difficult the task is on natural speech. The results obtained on synthetic speech are also very low compared to their textual counterpart.\footnote{\citet{merkx_frank_2019} report $\text{R@1}=27.5$ on a GRU-based model using characters as input.} For space reasons, and as the results on both Flickr8k and COCO show the same trend, we will focus in the following pages on analysing the results obtained on the Flickr8k data set. The results obtained on the COCO data set are reported in Appendix \ref{sec:appendix-coco}.

\begin{table*}[h!]
\resizebox{\linewidth}{!}{
\begin{tabular}{c|l:l|l:l|l:l|l:l||l:l|l:l|l:l|l:l}
 &
  \multicolumn{8}{c||}{{\color[HTML]{000000} Flickr8k  |  \keep condition}} &
  \multicolumn{8}{c}{{\color[HTML]{000000} Flickr8k |  \all condition}} \\ \hline\hline
GRU &
  \multicolumn{2}{c|}{Phones} &
  \multicolumn{2}{c|}{Syl.-Co.} &
  \multicolumn{2}{c|}{Syl.-Word} &
  \multicolumn{2}{c||}{Word} &
  \multicolumn{2}{c|}{Phones} &
  \multicolumn{2}{c|}{Syl.-Co.} &
  \multicolumn{2}{c|}{Syl.-Word} &
  \multicolumn{2}{c}{Word} \\ \cline{2-17} 
Pack. &
  \multicolumn{1}{c:}{T} &
  \multicolumn{1}{c|}{R} &
  \multicolumn{1}{c:}{T} &
  \multicolumn{1}{c|}{R} &
  \multicolumn{1}{c:}{T} &
  \multicolumn{1}{c|}{R} &
  \multicolumn{1}{c:}{T} &
  \multicolumn{1}{c||}{R} &
  \multicolumn{1}{c:}{T} &
  \multicolumn{1}{c|}{R} &
  \multicolumn{1}{c:}{T} &
  \multicolumn{1}{c|}{R} &
  \multicolumn{1}{c:}{T} &
  \multicolumn{1}{c|}{R} &
  \multicolumn{1}{c:}{T} &
  \multicolumn{1}{c}{R} \\ \hline
5 &
	% phones keep
	{\color[HTML]{000000} 3.6 }&% (True)
	{\color[HTML]{000000} \textbf{3.7} }&% (Random)
	% syllables-connected keep
	{\color[HTML]{000000} \textit{\textbf{3.6}} }&% (True)
	{\color[HTML]{000000} \textit{2.5} \textsuperscript{\textbf{--}}}&% (Random)
	% syllables-word keep
	{\color[HTML]{000000} \textbf{3.3} }&% (True)
	{\color[HTML]{000000} 3.0 }&% (Random)
	% words keep
	{\color[HTML]{000000} \textbf{3.2} }&% (True)
	{\color[HTML]{000000} \textbf{3.2} }&% (Random)
	%---------
	% phones all
	{\color[HTML]{000000} \textbf{4.0} }&% (True)
	{\color[HTML]{000000} 3.9 }&% (Random)
	% syllables-connected all
	{\color[HTML]{000000} \textbf{4.1} }&% (True)
	{\color[HTML]{000000} \textbf{4.1} }&% (Random)
	% syllables-word all
	{\color[HTML]{000000} \textbf{4.3} }&% (True)
	{\color[HTML]{000000} 3.9 }&% (Random)
	% words all
	{\color[HTML]{000000} 3.4 }&% (True)
	{\color[HTML]{000000} \textbf{4.2} }\\% (Random)
	%---------
4 &
	% phones keep
	{\color[HTML]{000000} \textbf{3.8} }&% (True)
	{\color[HTML]{000000} \textbf{3.8} }&% (Random)
	% syllables-connected keep
	{\color[HTML]{000000} \textbf{4.4} }&% (True)
	{\color[HTML]{000000} 3.5 }&% (Random)
	% syllables-word keep
	{\color[HTML]{000000} \textit{\textbf{3.9}} }&% (True)
	{\color[HTML]{000000} \textit{2.6} \textsuperscript{\textbf{--}}}&% (Random)
	% words keep
	{\color[HTML]{000000} \textit{\textbf{5.2}} \textsuperscript{\textbf{+}}}&% (True)
	{\color[HTML]{000000} \textit{2.5} \textsuperscript{\textbf{--}}}&% (Random)
	%---------
	% phones all
	{\color[HTML]{000000} 4.0 }&% (True)
	{\color[HTML]{000000} \textbf{4.4} }&% (Random)
	% syllables-connected all
	{\color[HTML]{000000} 3.9 }&% (True)
	{\color[HTML]{000000} \textbf{4.1} }&% (Random)
	% syllables-word all
	{\color[HTML]{000000} \textbf{4.3} }&% (True)
	{\color[HTML]{000000} 3.8 }&% (Random)
	% words all
	{\color[HTML]{000000} \textbf{4.5} }&% (True)
	{\color[HTML]{000000} \textbf{4.5} }\\% (Random)
	%---------
3 &
	% phones keep
	{\color[HTML]{000000} \textit{\textbf{4.9}} \textsuperscript{\textbf{+}}}&% (True)
	{\color[HTML]{000000} \textit{3.8} }&% (Random)
	% syllables-connected keep
	{\color[HTML]{000000} \textit{\textbf{4.5}} }&% (True)
	{\color[HTML]{000000} \textit{3.1} }&% (Random)
	% syllables-word keep
	{\color[HTML]{000000} \textit{\textbf{5.3}} \textsuperscript{\textbf{+}}}&% (True)
	{\color[HTML]{000000} \textit{3.1} }&% (Random)
	% words keep
	{\color[HTML]{000000} \textit{\textbf{4.9}} \textsuperscript{\textbf{+}}}&% (True)
	{\color[HTML]{000000} \textit{3.3} }&% (Random)
	%---------
	% phones all
	{\color[HTML]{000000} \textbf{4.5} }&% (True)
	{\color[HTML]{000000} 4.4 }&% (Random)
	% syllables-connected all
	{\color[HTML]{000000} \textbf{4.3} }&% (True)
	{\color[HTML]{000000} 4.2 }&% (Random)
	% syllables-word all
	{\color[HTML]{000000} \textbf{4.4} }&% (True)
	{\color[HTML]{000000} 4.2 }&% (Random)
	% words all
	{\color[HTML]{000000} \textbf{4.5} }&% (True)
	{\color[HTML]{000000} 3.8 }\\% (Random)
	%---------
2 &
	% phones keep
	{\color[HTML]{000000} \textbf{4.8} \textsuperscript{\textbf{+}}}&% (True)
	{\color[HTML]{000000} 3.9 }&% (Random)
	% syllables-connected keep
	{\color[HTML]{000000} \textit{\textbf{5.1}} \textsuperscript{\textbf{+}}}&% (True)
	{\color[HTML]{000000} \textit{3.6} }&% (Random)
	% syllables-word keep
	{\color[HTML]{000000} \textit{\textbf{4.8}} }&% (True)
	{\color[HTML]{000000} \textit{3.4} }&% (Random)
	% words keep
	{\color[HTML]{FF0000} \textit{\textbf{5.4}} \textsuperscript{\textbf{+}}}&% (True)
	{\color[HTML]{000000} \textit{3.4} }&% (Random)
	%---------
	% phones all
	{\color[HTML]{000000} \textbf{4.5} }&% (True)
	{\color[HTML]{000000} 3.8 }&% (Random)
	% syllables-connected all
	{\color[HTML]{000000} \textit{\textbf{4.8}} }&% (True)
	{\color[HTML]{000000} \textit{3.6} }&% (Random)
	% syllables-word all
	{\color[HTML]{000000} \textbf{4.4} }&% (True)
	{\color[HTML]{000000} 4.2 }&% (Random)
	% words all
	{\color[HTML]{000000} \textbf{4.7} }&% (True)
	{\color[HTML]{000000} 4.1 }\\% (Random)
	%---------
1 &
	% phones keep
	{\color[HTML]{000000} \textit{\textbf{4.8}} }&% (True)
	{\color[HTML]{000000} \textit{2.4} \textsuperscript{\textbf{--}}}&% (Random)
	% syllables-connected keep
	{\color[HTML]{000000} \textit{\textbf{3.4}} }&% (True)
	{\color[HTML]{000000} \textit{1.9} \textsuperscript{\textbf{--}}}&% (Random)
	% syllables-word keep
	{\color[HTML]{000000} \textit{\textbf{4.4}} }&% (True)
	{\color[HTML]{000000} \textit{2.0} \textsuperscript{\textbf{--}}}&% (Random)
	% words keep
	{\color[HTML]{000000} \textit{\textbf{3.9}} }&% (True)
	{\color[HTML]{000000} \textit{1.9} \textsuperscript{\textbf{--}}}&% (Random)
	%---------
	% phones all
	{\color[HTML]{000000} \textbf{4.3} }&% (True)
	{\color[HTML]{000000} 3.4 }&% (Random)
	% syllables-connected all
	{\color[HTML]{000000} \textbf{4.0} }&% (True)
	{\color[HTML]{000000} \textbf{4.0} }&% (Random)
	% syllables-word all
	{\color[HTML]{000000} \textbf{4.4} }&% (True)
	{\color[HTML]{000000} 4.3 }&% (Random)
	% words all
	{\color[HTML]{000000} \textit{\textbf{5.3}} \textsuperscript{\textbf{+}}}&% (True)
	{\color[HTML]{000000} \textit{4.1} }\\% (Random)
	%---------
\end{tabular}
}
\caption{Maximum R@1 (in \%) for each model trained on test set of the Flickr8k data set (models were selected based on the maximum R@1 on the validation set). ``T'' stands for \true (boundaries) and ``R'' stands for \random (boundaries). ``Syl-Co.'' and ``Syl-Word'' stand for ``Syllables-Connected'' and ``Syllables-Word'' respectively. Each line shows the results for when a specific recurrent layer is a \grupackager layer (see §\ref{sub-sec:grupack-position}). The $1^{st}$ layer is the lowest layer 
%(right after the 1D convolutions and acoustic vectors) 
and the $5^{th}$ the highest. 
%(right after the four preceding recurrent layers and before the attention mechanism). 
The highest R@1 in the table is shown in \textcolor{red}{red}. Best results between each \true and \random pair (columnwise) are shown in \textbf{bold}. ◌\textsuperscript{\textbf{+}} and ◌\textsuperscript{\textbf{--}} indicate that the results are statistically better (respectively worse) than the baseline. Results in \textit{italics} show statistical significance (two-sided Z-Test, p-value $< 1\mathrm{e}^{-2}$, see §\ref{subsec:evaluation}) between each \true and \random pair (columnwise). 
%A graphical representation of the results in the \random-\keep and \true-\keep condition is shown in Figure~\ref{fig:flickr-bar-plots}.
}
\label{tab:flickr-results}
\end{table*}

\textbf{\true/\random and \all/\keep Boundaries }
One of the questions our experiments aim at answering is whether introducing boundary information helps the network in solving its task or not. To do so, we first compare the difference between \true and \random boundaries. We notice different patterns depending on the position of the \grupackager layer and also depending on the \all and \keep conditions. 

We observe that in the \all condition the results between \true and \random boundaries are overall not statistically different from one another, and are not significantly better or worse from the baseline results. There is only one case where such differences are statistically significant: for the $1^{st}$ layer when using word segments. However, in the \keep condition, we observe a strong difference between \true and \random boundaries across all boundary types and across most of the layers. Overall, in the \keep condition, models trained with \true boundaries have statistically different results from models trained with \random boundaries. Also, in such settings, the results obtained are generally statistically better than the baseline, while in the \random-\keep condition the results are statistically worse than the baseline.

These results show that there is overall no difference between using \true or \random boundaries in the \all condition (except for one layer), hence showing that boundary information is not used effectively by the network. In contrast, the difference between \true and \random in the \keep condition shows that boundary information is effectively used by the network. Using random boundaries which do not delimit meaningful linguistic units really hurts the performance of the network, especially in the \keep condition as most of the vectors are removed. In the \all condition, using \true or \random boundaries yields results close to that of the \baseline, suggesting boundary information might act as noise and help the network regularise.
Thus, as expected, the network was effectively constrained to learn better representations in the \keep condition.  We believe it is the case because in the \all condition, boundary information is diluted among the neighbouring vectors while this is not the case in the \keep condition, as each segment is represented by a single vector.

\textbf{Phones, Syllables, or Words}
From now on, we will focus on the results obtained in the \keep condition, as the \all condition brings only slight improvement over the \baseline condition. In our experiments we used four different type of segments corresponding to different type of linguistic units: phones, syllables-connected, syllables-word, and words. These different type of segments vary in \emph{length} (words and syllables are longer than phones), \emph{quantity} (there are more phones and syllables than words), and \emph{intrinsic linguistic information}: phones only show which are the basic acoustic units of the language, while word segments represent meaningful units, and syllables-word and syllables-connected are a higher form of acoustic unit that may contain morphemic information. Given the task the network is trained for (speech-image retrieval), we do not expect these different units to perform equally well. Indeed, as this task implies mapping an image vector describing which objects are present in a picture and a spoken description of an image, we expect word-like segments (or segments that preserve word boundaries and that bear a substantial amount of semantic information) to perform better. 

This is in fact what we observe in practice: word units obtain statistically better results ($R@1=5.4$) than the baseline ($+1.1$pp). Syllables-word also bring significant improvement ($R@1=5.3$), however, slightly less than when using word units.
It should be noted that syllables-connected segments also obtain  statistically significant improvement over the baseline (\grupackager[2]) despite not preserving all the word boundaries. However, these results are slightly worse than the syllables-word and word segments, suggesting that preserving word boundaries is a property that helps the network. It appears that the size of a segment is also an important parameter. Indeed, phone segments (naturally) preserve word boundaries, but of course naturally lack the internal cohesion of a morpheme or a word as nothing links two adjacent phonemes together. Thus, it seems that segments that preserve meaning (such as words) or from which meaning can be more easily recomposed (syllable) may facilitate the network's task. The fact that syllable-like segments perform as well as word segments might only be an artefact of using English where a high proportion of word is monosyllabic.\footnote{\citet{jespersen_monosyllabism} estimates that at least 8,000 commonly frequent words are monosyllabic in English.} Working on a language such as Japanese where the syllable-to-morpheme ratio is
higher would be a future line of work that would enable to test this hypothesis.
\begin{figure*}[h!]
	\begin{minipage}{.32\linewidth}
		\centering
		\subfloat[]{\label{tsne-baseline}\includegraphics[width=1.05\textwidth]{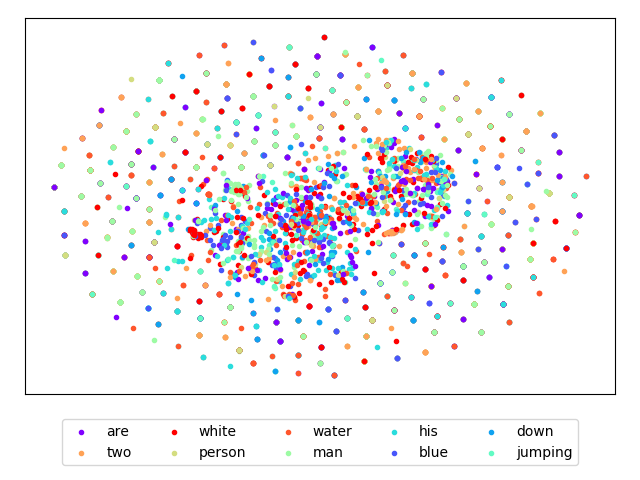}}
	\end{minipage}
	\begin{minipage}{.32\linewidth}
		\centering
		\subfloat[]{\label{tsne-all}\includegraphics[width=1.05\textwidth]{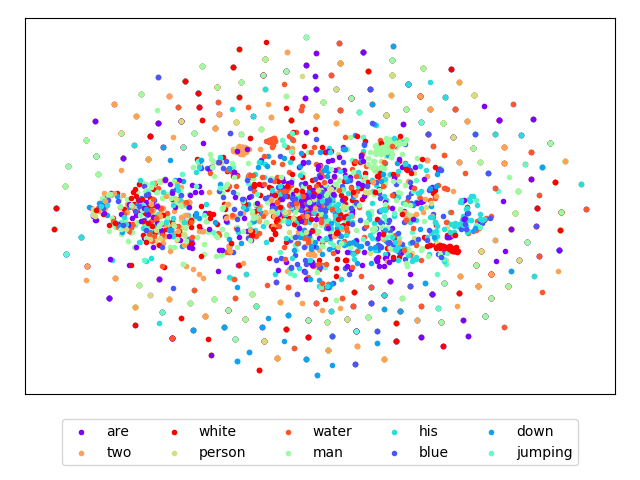}}
	\end{minipage}
	\begin{minipage}{.32\linewidth}
		\centering
		\subfloat[]{\label{tsne-keep}\includegraphics[width=1.05\textwidth]{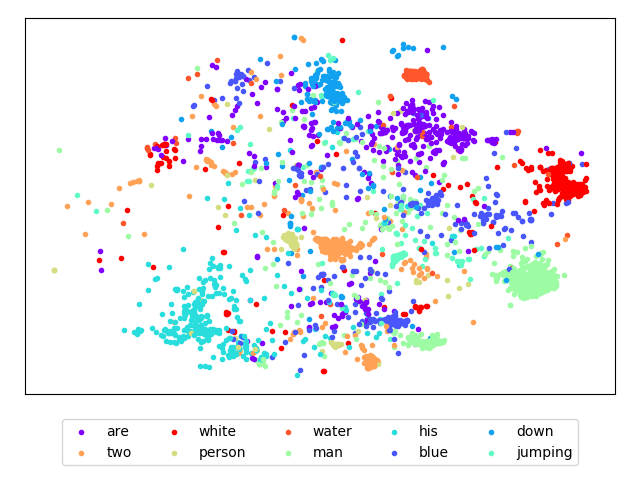}}
	\end{minipage}
	\caption{t-SNE projections of the final vector of different occurrences of eigth randomly selected words (Flick8k) in the \baseline~condition~(\ref{tsne-baseline}), in the \all~condition~(\ref{tsne-all}), and in the \keep~condition~(\ref{tsne-keep}). %In all cases, the vectors projected correspond to the vectors $h_2, h_5 \text{, and }h_6$ of Figure~\ref{fig:gru}. 
	Plot \ref{tsne-baseline} shows that the representation learnt in the \baseline and \all conditions are not word-based as the final vectors of different occurrences of the same word do not cluster together. 
	%In the \all condition, the model overall fails to learn similar representations for the same occurrences of the same word (except for one word: ``man''), despite being supplied with word boundaries. 
	In the \keep condition, the model succeeded in learning similar representations for different occurrences of the same word as words cluster together.}
	\label{fig:tsne}
\end{figure*}

\textbf{\grupackager Layer Position}
We introduced boundary information at different levels of our architecture in order to better understand at which layer it is the most useful to add such information.
Our results clearly show that introducing boundary information at  different layers has a clear impact on the results: using such information at the first or the fifth layer is useless, as we notice it either yields similar results to the baseline or worsens the results regardless of the type of boundary used (\grupackager[5]). When using syllable-word segments the best results are obtained when introducing the information at \grupackager[3], and at \grupackager[2] when using word segments. 
Word-like segments seem to be the most robust representation to be used as they yield significantly better results at three different layers (\grupackager[2,3,4]). We also notice that phone segments bring no improvement over the baseline at \grupackager[4,5] showing that these layers do not handle phone like information.
All in all, these results are in line with that of \citet{Chrupala2017} who found that the intermediate representations of the fifth layer is the less informative in predicting word presence, while lower layers encode this information better. This confirms that the middle layers of our architecture deal with lexical units whereas the fifth layer encodes information that disregards that type of information.

\subsection{Segmentation as a Means for Compression}
\label{sec:seg-comp}
Recall that in the \keep condition, only the last vector comprising a segment is kept while the other vectors are discarded. This can be interpreted as a form of ``guided'' subsampling, as usually subsampling does not take into consideration linguistic factors. To understand how much information is kept between the input and the output of a \grupackager layer in the \keep condition, we compute an average compression rate (in~\%) for each of the segment types for Flickr8k. The results are the following: phones~=~90.57\%, syllables-connected~=~93.41\%, syllables-word~=~94.36\%, and words~=~94.90\%.\footnote{Note that the compression rate for syllables-word and words is very close, suggesting there is a significant overlap between syllables-word units and word units.}
When we re-analyse our results in light of this information, it appears we can remove a large part of the original input (up to 94.90\% if using word segments) while conserving or increasing the original R@1. It is not simply the effect of subsampling that helps, but subsampling with \emph{meaningful} linguistic units. The effect of informed subsampling is striking when we compare R@1 for \random-\keep, which are always below the \baseline, while \true-\keep are on a par with the \baseline or better.
A counter-intuitive finding of our experiments is that it is better to subsample early on (in the first layers) and thus remove most of the information early on than later on. Subsampling with word segments in \grupackager[2] (and thus only keeping 5.1\% of the original amount of information for the subsequent layers) yields better results than subsampling with the same resolution at \grupackager[5].

\begin{table*}[!htbp]
\resizebox{\textwidth}{!}{  
    \centering
\begin{tabular}{|c|ccccc|cccc|ccc|cc|}
\hline
\textbf{Architecture} & \multicolumn{5}{c|}{\textbf{5 layers}}                                                                                                                                                    & \multicolumn{4}{c|}{\textbf{4 layers}}                                                                                           & \multicolumn{3}{c|}{\textbf{3 layers}}                                                         & \multicolumn{2}{c|}{\textbf{2 layers}}                       \\ \hline\hline
\diagbox{$1^{st}\text{\grupackager}$}
{$2^{nd}\text{\grupackager}$}     & \multicolumn{1}{c|}{\textbf{1}} & \multicolumn{1}{c|}{\textbf{2}} & \multicolumn{1}{c|}{\textbf{3}}                          & \multicolumn{1}{c|}{\textbf{4}} & \textbf{5}               & \multicolumn{1}{c|}{\textbf{1}} & \multicolumn{1}{c|}{\textbf{2}}   & \multicolumn{1}{c|}{\textbf{3}} & \textbf{4}               & \multicolumn{1}{c|}{\textbf{1}} & \multicolumn{1}{c|}{\textbf{2}}   & \textbf{3}               & \multicolumn{1}{c|}{\textbf{1}}   & \textbf{2}               \\ \hline
\textbf{1}                       & \multicolumn{1}{c|}{\cellcolor[HTML]{C0C0C0}} & \multicolumn{1}{c|}{\cellcolor[HTML]{FFFFFF}7.7} & \multicolumn{1}{c|}{\cellcolor[HTML]{FFFFFF}7.7}                                 & \multicolumn{1}{c|}{\cellcolor[HTML]{FFFFFF}7.3} & \cellcolor[HTML]{FFFFFF}3.9 & \multicolumn{1}{c|}{\cellcolor[HTML]{C0C0C0}{\color[HTML]{333333} }} & \multicolumn{1}{c|}{\cellcolor[HTML]{FFFFFF}{\color[HTML]{333333} 7.6}}       & \multicolumn{1}{c|}{\cellcolor[HTML]{FFFFFF}{\color[HTML]{333333} 7.9}}          & \cellcolor[HTML]{FFFFFF}{\color[HTML]{333333} 5.7} & \multicolumn{1}{c|}{\cellcolor[HTML]{C0C0C0}{\color[HTML]{333333} }} & \multicolumn{1}{c|}{\cellcolor[HTML]{FFFFFF}{\color[HTML]{333333} \textbf{8.1}}} & \cellcolor[HTML]{FFFFFF}{\color[HTML]{333333} 5.3} & \multicolumn{1}{c|}{\cellcolor[HTML]{C0C0C0}{\color[HTML]{333333} \textbf{}}} & \cellcolor[HTML]{FFFFFF}{\color[HTML]{333333} \textbf{6.4}} \\ \cline{1-1} \cline{3-6} \cline{8-10} \cline{12-13} \cline{15-15} 
\textbf{2}                       & \cellcolor[HTML]{C0C0C0}                      & \multicolumn{1}{c|}{\cellcolor[HTML]{C0C0C0}}    & \multicolumn{1}{c|}{\cellcolor[HTML]{FFFFFF}{\color[HTML]{FE0000} \textbf{8.2}}} & \multicolumn{1}{c|}{\cellcolor[HTML]{FFFFFF}7.6} & \cellcolor[HTML]{FFFFFF}5.8 & \cellcolor[HTML]{C0C0C0}{\color[HTML]{333333} }                      & \multicolumn{1}{c|}{\cellcolor[HTML]{C0C0C0}{\color[HTML]{333333} \textbf{}}} & \multicolumn{1}{c|}{\cellcolor[HTML]{FFFFFF}{\color[HTML]{333333} \textbf{8.1}}} & \cellcolor[HTML]{FFFFFF}{\color[HTML]{333333} 6.3} & \cellcolor[HTML]{C0C0C0}{\color[HTML]{333333} }                      & \multicolumn{1}{c|}{\cellcolor[HTML]{C0C0C0}{\color[HTML]{333333} }}             & \cellcolor[HTML]{FFFFFF}{\color[HTML]{333333} 7.3} & \cellcolor[HTML]{C0C0C0}{\color[HTML]{333333} }                               & \cellcolor[HTML]{C0C0C0}{\color[HTML]{333333} }             \\ \cline{1-1} \cline{4-6} \cline{9-10} \cline{13-13}
\textbf{3}                       & \cellcolor[HTML]{C0C0C0}                      & \cellcolor[HTML]{C0C0C0}                         & \multicolumn{1}{c|}{\cellcolor[HTML]{C0C0C0}}                                    & \multicolumn{1}{c|}{\cellcolor[HTML]{FFFFFF}7.1} & \cellcolor[HTML]{FFFFFF}6.5 & \cellcolor[HTML]{C0C0C0}{\color[HTML]{333333} }                      & \cellcolor[HTML]{C0C0C0}{\color[HTML]{333333} }                               & \multicolumn{1}{c|}{\cellcolor[HTML]{C0C0C0}{\color[HTML]{333333} }}             & \cellcolor[HTML]{FFFFFF}{\color[HTML]{333333} 6.7} & \cellcolor[HTML]{C0C0C0}{\color[HTML]{333333} }                      & \cellcolor[HTML]{C0C0C0}{\color[HTML]{333333} }                                  & \cellcolor[HTML]{C0C0C0}{\color[HTML]{333333} }    & \cellcolor[HTML]{C0C0C0}{\color[HTML]{333333} }                               & \cellcolor[HTML]{C0C0C0}{\color[HTML]{333333} }             \\ \cline{1-1} \cline{5-6} \cline{10-10}
\textbf{4}                       & \cellcolor[HTML]{C0C0C0}                      & \cellcolor[HTML]{C0C0C0}                         & \cellcolor[HTML]{C0C0C0}                                                         & \multicolumn{1}{c|}{\cellcolor[HTML]{C0C0C0}}    & \cellcolor[HTML]{FFFFFF}6.1 & \cellcolor[HTML]{C0C0C0}{\color[HTML]{333333} }                      & \cellcolor[HTML]{C0C0C0}{\color[HTML]{333333} }                               & \cellcolor[HTML]{C0C0C0}{\color[HTML]{333333} }                                  & \cellcolor[HTML]{C0C0C0}{\color[HTML]{333333} }    & \cellcolor[HTML]{C0C0C0}{\color[HTML]{333333} }                      & \cellcolor[HTML]{C0C0C0}{\color[HTML]{333333} }                                  & \cellcolor[HTML]{C0C0C0}{\color[HTML]{333333} }    & \cellcolor[HTML]{C0C0C0}{\color[HTML]{333333} }                               & \cellcolor[HTML]{C0C0C0}{\color[HTML]{333333} }             \\ \cline{1-1} \cline{6-6}
\textbf{5}                       & \cellcolor[HTML]{C0C0C0}                      & \cellcolor[HTML]{C0C0C0}                         & \cellcolor[HTML]{C0C0C0}                                                         & \cellcolor[HTML]{C0C0C0}                         & \cellcolor[HTML]{C0C0C0}    & \cellcolor[HTML]{C0C0C0}{\color[HTML]{333333} }                      & \cellcolor[HTML]{C0C0C0}{\color[HTML]{333333} }                               & \cellcolor[HTML]{C0C0C0}{\color[HTML]{333333} }                                  & \cellcolor[HTML]{C0C0C0}{\color[HTML]{333333} }    & \cellcolor[HTML]{C0C0C0}{\color[HTML]{333333} }                      & \cellcolor[HTML]{C0C0C0}{\color[HTML]{333333} }                                  & \cellcolor[HTML]{C0C0C0}{\color[HTML]{333333} }    & \cellcolor[HTML]{C0C0C0}{\color[HTML]{333333} }                               & \cellcolor[HTML]{C0C0C0}{\color[HTML]{333333} }             \\ \hline
\hline\textbf{Baseline (No \grupackager)}     & \multicolumn{5}{c|}{4.3}                                                                                                                                                                  & \multicolumn{4}{c|}{4.4}                                                                                                         & \multicolumn{3}{c|}{3.4}                                                                       & \multicolumn{2}{c|}{3.5}                                     \\ \hline
\end{tabular}
}
    \caption{R@1 obtained on the test set of the Flickr8k data set with a hierarchical architecture consisting of two \grupackager layers using phone and word segments (models were selected based on the maximum R@1 on the validation set). Best score overall is shown in \textcolor{red}{red}. Best score (layer-wise) is shown in \textbf{bold}. Greyed out cell signal impossible configurations. We also indicate R@1 obtained on a baseline architecture without any \grupackager.
    }
    \label{tab:results-hier-phones-words}
\end{table*}
\section{Towards Hierarchical Segmentation}
\label{sec:hierarchy}
In our current approach, only one out of the five recurrent layers is a \grupackager layer, which handles only one type of segment. However, we can stack as many \grupackager as desired, provided they are supplied with boundary information. Stacking such layers enables us to not only integrate boundary information, but also introduce structure, where one layer handles one type of segment (e.g. phone) and the following \grupackager layer handles another type of segment, that is hierarchically above the preceding (e.g. syllable, or word).\footnote{Note that it could also be possible to use larger units, such as chunks.} \citet{harwath_hierachical} explored such hierarchical architecture using a CNN-based model that incorporated vector quantisation layers and found that it improves R@k. Our work thus attempts to verify if it is also the case for an RNN-based model.

\textbf{Phones and Words}: We first explore hierarchical segmentation with phones and words on the Flickr8k data set.\footnote{We also explored two other hierarchical architectures that use phones and syllables-word, and syllables-word and words. The results are reported in Appendix~\ref{sec:appendix-hier} in Table~\ref{tab:results-hier-phone-syllable} and Table~\ref{tab:results-hier-syllable-word}.}
We only consider the \keep condition as it yields better results than the \all condition. We vary the position of the \grupackager layers as well as the number of layers (from 2 to 5) and test all possible positions with two \grupackager layers. For each configuration, the lowest \grupackager will receive phone boundary information, and the next \grupackager layer will receive word boundary information. Note that such configuration results in a double sequence reduction. Indeed, after the first \grupackager layer, they are only as many output vectors as there are phones, and in the second, the resulting phone vectors are recomposed together to form words, resulting in as many output vectors as there are words. Results are shown in Table~\ref{tab:results-hier-phones-words}. Training an architecture with two \grupackager layers, each handling two different types of segments results in much better R@1 than the baseline ($+3.9$pp) and than a single-\grupackager-layered architecture ($+2.8$pp), thus showing that introducing hierarchy is beneficial. Results also confirm that the layer 2 and 3 of our architecture are those that benefit the most from adding linguistic information, and confirm the fact that the upper layers (such as the fifth) do not take as much advantage of this information as the lower layers. Introducing structure allowed us to remove two recurrent layers without a big loss of performance ($R@1=8.1$ for a three-layered architecture with two \grupackager layers) while the baseline architecture with only three layers performs poorly.

\textbf{Phones, Syllables, and Words}: We also explore an architecture with 3 \grupackager layers, to which we provide phone, syllable-word and word boundaries. As in our previous experiments, we vary the number of layers (from 3 to 5), and test all possible configurations. The results of this experiment are presented in Table \ref{tab:results-hier-phone-syllable-word}. 
We notice that the best result obtained with this architecture is far superior to the best result of a single-layered architecture ($R@1=9.6, +4.2$pp), but also superior to the best result of a double-layered architecture ($+1.4$pp over the phone-word architecture). Our best results are obtained by a five-layered architecture with \grupackager in position 1, 3 and 4. However, we notice that the four-layered architecture obtains more consistent results across all layers, the maximum result being only $-0.3$pp away from best five-layered architecture. We also notice that the 3 layered architecture obtains a very high R@1 of $8.0$ which is about two times over the baseline results. 

Our results show that the more structure we introduce into the network, the better it performs. Additionally, introducing structure enables us to reduce the number of layers (and hence the number of computations) while increasing the performances compared to the baseline. Overall, it is better to use boundary information in coordination in a hierarchical structure than using them in isolation.

\begin{table}[h!]
	\centering
	\resizebox{\columnwidth}{!}{
	\begin{tabular}{|c|c|c|c|}
		\hline
		\diagbox{\grupackager}{Architecture} & \textbf{5 layers} & \textbf{4 layers}                    & \textbf{3 layers}        \\ \hline
		1 + 2 + 3                                & 8.5               & { \textbf{9.3} } & 8.0                      \\ \hline
		1 + 2 + 4                                & 8.1               & 8.6 & \cellcolor[HTML]{C0C0C0}                      \\ \cline{1-2}
		1 + 2 + 5                                & 7.8               & \cellcolor[HTML]{C0C0C0} & \cellcolor[HTML]{C0C0C0}                      \\ \cline{1-2}
		1 + 3 + 4                                & \color[HTML]{FE0000}\textbf{9.6}               & 8.4 & \cellcolor[HTML]{C0C0C0}                      \\ \cline{1-3}
		1 + 3 + 5                                & 7.9               & \cellcolor[HTML]{C0C0C0} & \cellcolor[HTML]{C0C0C0}                      \\ \cline{1-2}
		1 + 4 + 5                                & 7.8               & \cellcolor[HTML]{C0C0C0} & \cellcolor[HTML]{C0C0C0}                      \\ \cline{1-2}
		2 + 3 + 4                                & 8.8               & 8.3                                  & \cellcolor[HTML]{C0C0C0} \\ \cline{1-4}
		2 + 3 + 5                               & 8.5               & \cellcolor[HTML]{C0C0C0} & \cellcolor[HTML]{C0C0C0}                     \\ \cline{1-2}
		2 + 4 + 5                                & 8.3               & \cellcolor[HTML]{C0C0C0} & \cellcolor[HTML]{C0C0C0}                      \\ \cline{1-2}
		3 + 4 + 5                                & 7.8               & \cellcolor[HTML]{C0C0C0}             & \cellcolor[HTML]{C0C0C0} \\ \hline
	\end{tabular}
	}
	\caption{R@1 obtained on the test set of the Flickr8k data set with a hierarchical architecture consisting of three \grupackager layers using phone, syllable-word and word segments (models were selected based on the maximum R@1 on the validation set). 
	%The same naming conventions of Table \ref{tab:results-hier-phones-words} are used for this table
	}
	\label{tab:results-hier-phone-syllable-word}
\end{table}

\section{Discussion and Future Work}
\label{sec:discussion}
The goal of our experiments is to see if segmenting speech in sub-units is beneficial, and if so, which units maximise the performance. It is indeed the case that segmenting speech into sub-units helps. As to which segment obtains the best performance we observe mixed results. Indeed, word segmentation yields better results than phone segmentation, but we do also observe that syllable-like segmentation also gives results that are in the same ballpark as word segmentation. Nevertheless, word segmentation seems to be a \emph{more robust} representation compared to syllable as such word segments consistently yield better results at various levels of our architecture.

Another finding of our experiments which we believe is important is that one cannot subsample speech without taking into account its linguistic nature. Indeed, random subsampling might yield results on a par with the baseline, but this might only be a regularisation effect. \emph{Linguistically informed subsampling} (\keep condition) yields however much better results and should be favoured. 

Regarding the question of why textual approaches perform better than spoken approaches, we conclude that the fact that tokens stand for full semantic units plays little in their performance. The fact that text-based models use segmented input (either tokens or characters) also seems to play little in the final performance, otherwise we should have observed better results as our input was also segmented. What seems most crucial is that the representation of a token never changes whereas speech exhibits lots of variation, as no word is pronounced exactly in the same fashion when uttered. Our approach helped the network in building more consistent representations for the same word (especially in the \keep condition, see Figure~\ref{fig:tsne}), even though it did not succeed for every word.  Consistent representation across various occurrences seems to be the most important factor. 

Finally, our experiments allowed us to observe that, such as for humans, the use of large units, such as words, is indeed the most efficient solution to learn a reliable speech-to-image mapping. Indeed, even if phone and syllable-like units yield non trivial results, they are less robust than word segments. Our \grupackager setting also allowed us to simply introduce hierarchy in a neural network by simply stacking \grupackager layers and providing different boundary information to each of them. Our experiments allowed us to confirm the results obtained by \citet{harwath_hierachical} on a CNN-based VGS model, stating that introducing hierarchical structures proves beneficial overall even for shallower architectures. Interestingly, our hierarchical experiments allowed us to notice that using segment boundaries in isolation only brings slight improvements. It is only when different levels are combined (phones and words, or phones, syllables and words) that the performance of the network reaches its peak.

The future lines of work we imagine consist in \emph{learning} where the boundaries are located instead of
supplying boundary information to the network at training and testing time.
We could indeed use ACT recurrent cells \cite{kreutzer_act} or an architecture such as \cite{audio_word2vec} that would dynamically and unsupervisedly learn how to segment the input signal into sub-units. The additional advantage of such methods is that they make no presupposition on the form/size of the segments, and consequently on what a good segment should or should not be, but lets the network find the optimal solution. Finally, we plan to also introduce syntactic information and integrate chunk boundaries and measure the impact of syntactical grouping of spoken units.

\section*{Acknowledgments}
This work was supported by grants from NeuroCoG IDEX UGA as part of the ``Investissements d'avenir" program (ANR-15-IDEX-02).

\bibliographystyle{acl_natbib}
\bibliography{main}

%
%   Supplemental Material
%
\clearpage
\pagebreak
\newpage
\begin{table}[b]\setlength{\hfuzz}{1.1\columnwidth}
\begin{minipage}{\textwidth}
%\begin{minipage}{\textwidth}

\resizebox{1\textwidth}{!}{
\begin{tabular}{c|l:l|l:l|l:l|l:l||l:l|l:l|l:l|l:l}
      & \multicolumn{8}{c||}{COCO | \keep condition}  &    \multicolumn{8}{c}{COCO | \all condition}                                 \\ \hline\hline
GRU &
  \multicolumn{2}{c|}{Phones} &
  \multicolumn{2}{c|}{Syl.-Co.} &
  \multicolumn{2}{c|}{Syl.-Word} &
  \multicolumn{2}{c||}{Word} &
  \multicolumn{2}{c|}{Phones} &
  \multicolumn{2}{c|}{Syl.-Co.} &
  \multicolumn{2}{c|}{Syl.-Word} &
  \multicolumn{2}{c}{Word} \\ \cline{2-17} 
Pack. &
  \multicolumn{1}{c:}{T} &
  \multicolumn{1}{c|}{R} &
  \multicolumn{1}{c:}{T} &
  \multicolumn{1}{c|}{R} &
  \multicolumn{1}{c:}{T} &
  \multicolumn{1}{c|}{R} &
  \multicolumn{1}{c:}{T} &
  \multicolumn{1}{c||}{R} &
  \multicolumn{1}{c:}{T} &
  \multicolumn{1}{c|}{R} &
  \multicolumn{1}{c:}{T} &
  \multicolumn{1}{c|}{R} &
  \multicolumn{1}{c:}{T} &
  \multicolumn{1}{c|}{R} &
  \multicolumn{1}{c:}{T} &
  \multicolumn{1}{c}{R} \\ \hline
5 &
	% phones keep
	{\color[HTML]{000000} 9.4 }&% (True)
	{\color[HTML]{000000} \textbf{9.6} }&% (Random)
	% syllables-connected keep
	{\color[HTML]{000000} \textbf{9.1} }&% (True)
	{\color[HTML]{000000} \textbf{9.1} }&% (Random)
	% syllables-word keep
	{\color[HTML]{000000} \textbf{9.6} }&% (True)
	{\color[HTML]{000000} 9.1 }&% (Random)
	% words keep
	{\color[HTML]{000000} \textit{\textbf{9.4}} }&% (True)
	{\color[HTML]{000000} \textit{8.7} }&% (Random)
	%---------
	% phones all
	{\color[HTML]{000000} \textbf{9.7} \textsuperscript{\textbf{+}}}&% (True)
	{\color[HTML]{000000} 9.4 }&% (Random)
	% syllables-connected all
	{\color[HTML]{000000} 9.1 }&% (True)
	{\color[HTML]{000000} \textbf{9.5} }&% (Random)
	% syllables-word all
	{\color[HTML]{000000} \textbf{9.5} }&% (True)
	{\color[HTML]{000000} 9.4 }&% (Random)
	% words all
	{\color[HTML]{000000} 9.3 }&% (True)
	{\color[HTML]{000000} \textbf{9.5} }\\% (Random)
	%---------
4 &
	% phones keep
	{\color[HTML]{000000} 10.0 \textsuperscript{\textbf{+}}}&% (True)
	{\color[HTML]{000000}\textbf{ 10.5} \textsuperscript{\textbf{+}}}&% (Random)
	% syllables-connected keep
	{\color[HTML]{000000} \textbf{10.2} \textsuperscript{\textbf{+}}}&% (True)
	{\color[HTML]{000000} 9.6 }&% (Random)
	% syllables-word keep
	{\color[HTML]{000000} \textbf{10.4} \textsuperscript{\textbf{+}}}&% (True)
	{\color[HTML]{000000} 9.9 \textsuperscript{\textbf{+}}}&% (Random)
	% words keep
	{\color[HTML]{000000} \textit{\textbf{10.6}} \textsuperscript{\textbf{+}}}&% (True)
	{\color[HTML]{000000} \textit{9.5} }&% (Random)
	%---------
	% phones all
	{\color[HTML]{000000} \textbf{9.3} }&% (True)
	{\color[HTML]{000000} 9.2 }&% (Random)
	% syllables-connected all
	{\color[HTML]{000000} \textbf{9.4} }&% (True)
	{\color[HTML]{000000} 9.1 }&% (Random)
	% syllables-word all
	{\color[HTML]{000000} \textbf{9.6} }&% (True)
	{\color[HTML]{000000} 9.2 }&% (Random)
	% words all
	{\color[HTML]{000000} 9.0 }&% (True)
	{\color[HTML]{000000}\textbf{9.4} }\\% (Random)
	%---------
3 &
	% phones keep
	{\color[HTML]{000000} \textbf{10.5} \textsuperscript{\textbf{+}}}&% (True)
	{\color[HTML]{000000} 10.1 \textsuperscript{\textbf{+}}}&% (Random)
	% syllables-connected keep
	{\color[HTML]{000000} \textbf{10.4} \textsuperscript{\textbf{+}}}&% (True)
	{\color[HTML]{000000} 9.8 \textsuperscript{\textbf{+}}}&% (Random)
	% syllables-word keep
	{\color[HTML]{000000} \textbf{10.5} \textsuperscript{\textbf{+}}}&% (True)
	{\color[HTML]{000000} 10.1 \textsuperscript{\textbf{+}}}&% (Random)
	% words keep
	{\color[HTML]{000000} \textit{\textbf{11.0}} \textsuperscript{\textbf{+}}}&% (True)
	{\color[HTML]{000000} \textit{9.7} }&% (Random)
	%---------
	% phones all
	{\color[HTML]{000000} \textbf{9.5} }&% (True)
	{\color[HTML]{000000} 9.2 }&% (Random)
	% syllables-connected all
	{\color[HTML]{000000} \textbf{9.2} }&% (True)
	{\color[HTML]{000000} 9.1 }&% (Random)
	% syllables-word all
	{\color[HTML]{000000} \textbf{9.4} }&% (True)
	{\color[HTML]{000000} 9.1 }&% (Random)
	% words all
	{\color[HTML]{000000} \textbf{9.4} }&% (True)
	{\color[HTML]{000000} 9.2 }\\% (Random)
	%---------
2 &
	% phones keep
	{\color[HTML]{000000} \textit{\textbf{10.7}} \textsuperscript{\textbf{+}}}&% (True)
	{\color[HTML]{000000} \textit{9.8} \textsuperscript{\textbf{+}}}&% (Random)
	% syllables-connected keep
	{\color[HTML]{000000} \textit{\textbf{10.5}} \textsuperscript{\textbf{+}}}&% (True)
	{\color[HTML]{000000} \textit{9.4} }&% (Random)
	% syllables-word keep
	{\color[HTML]{000000} \textit{\textbf{10.9}} \textsuperscript{\textbf{+}}}&% (True)
	{\color[HTML]{000000} \textit{9.3} }&% (Random)
	% words keep
	{\color[HTML]{FF0000} \textit{\textbf{11.3}} \textsuperscript{\textbf{+}}}&% (True)
	{\color[HTML]{000000} \textit{8.8} }&% (Random)
	%---------
	% phones all
	{\color[HTML]{000000} \textbf{9.4} }&% (True)
	{\color[HTML]{000000} 8.9 }&% (Random)
	% syllables-connected all
	{\color[HTML]{000000} \textbf{9.7} \textsuperscript{\textbf{+}}}&% (True)
	{\color[HTML]{000000} 9.1 }&% (Random)
	% syllables-word all
	{\color[HTML]{000000} \textbf{9.6} }&% (True)
	{\color[HTML]{000000} 9.2 }&% (Random)
	% words all
	{\color[HTML]{000000} \textit{\textbf{9.7}} \textsuperscript{\textbf{+}}}&% (True)
	{\color[HTML]{000000} \textit{8.9} }\\% (Random)
	%---------
1 &
	% phones keep
	{\color[HTML]{000000} \textit{\textbf{10.1}} \textsuperscript{\textbf{+}}}&% (True)
	{\color[HTML]{000000} \textit{7.9} \textsuperscript{\textbf{--}}}&% (Random)
	% syllables-connected keep
	{\color[HTML]{000000} \textit{\textbf{9.7}} }&% (True)
	{\color[HTML]{000000} \textit{7.1} \textsuperscript{\textbf{--}}}&% (Random)
	% syllables-word keep
	{\color[HTML]{000000} \textit{\textbf{10.2}} \textsuperscript{\textbf{+}}}&% (True)
	{\color[HTML]{000000} \textit{7.0} \textsuperscript{\textbf{--}}}&% (Random)
	% words keep
	{\color[HTML]{000000} \textit{\textbf{10.3}} \textsuperscript{\textbf{+}}}&% (True)
	{\color[HTML]{000000} \textit{7.0} \textsuperscript{\textbf{--}}}&% (Random)
	%---------
	% phones all
	{\color[HTML]{000000} \textbf{9.8} \textsuperscript{\textbf{+}}}&% (True)
	{\color[HTML]{000000} 9.4 }&% (Random)
	% syllables-connected all
	{\color[HTML]{000000} \textbf{9.6} }&% (True)
	{\color[HTML]{000000} 9.4 }&% (Random)
	% syllables-word all
	{\color[HTML]{000000} \textit{\textbf{10.0}} \textsuperscript{\textbf{+}}}&% (True)
	{\color[HTML]{000000} \textit{9.1} }&% (Random)
	% words all
	{\color[HTML]{000000} \textbf{9.5} }&% (True)
	{\color[HTML]{000000} 9.1 }\\% (Random)
	%---------

\end{tabular}
}
\captionof{table}{Maximum R@1 (in \%) for each model trained on the COCO data set. The same naming conventions of Table~\ref{tab:flickr-results} are used for this table.}
\label{tab:coco-results}

\end{minipage}
\end{table}

\appendix
\section{Results COCO}
\label{sec:appendix-coco}

\subsection{\all and \keep}

As for Flickr8k, the results obtained in the \all and the \keep condition show that the \all condition brings little improvement over the baseline results. However, we may observe a different pattern in the \random-\keep condition: whereas for Flickr8k the results are significatively worse in this
condition, some results are significatively better than the baseline for COCO (see for example the results obtained at \grupackager[3,4] with phones, syllables-connected and syllables word). We explain this by the fact that, contrary to Flickr8k that used real human speech, COCO uses synthetic speech with only one voice and hence, has very low intra-speaker variation. Thus, even though we randomly subsample the input, as there is very little intra-speaker variation, the network is much more likely to figure out from which units the subsampled vector came from. Thus, randomly subsampling the spoken input acts as a form of regularisation for the network such as dropout.

\subsection{Phones, Syllables, or Words}
We also observe that the best results are obtained when we use word segments (\grupackager[2]). As for Flickr8k, word units yield more consistent results over most of the layers (\grupackager[1,2,3,4]) suggesting that word-like segmentation is the adequate segmentation to be used for our task. We also notice that syllables word, that preserve word boundaries, obtain results close to that of word segments. As for Flickr8k, syllables-connected overall yield worse results that phones or syllables-word, once again showing that preserving word boundaries seems to be important.

\subsection{\grupackager Layer Position}
Results on the COCO data set also show that the worse results are obtained when boundary information is provided at the last layer (\grupackager[5]) showing that this layer is not concerned with form anymore, but with semantics. The best results (be it with phones, syllables-connected, syllables-word, or word) are obtained at the second layer. Results then decrease in the upper layer.

\section{Hierarchical Segmentation}
\label{sec:appendix-hier}
The results obtained with hierarchical models that use phones and syllables-word, and syllables-word and words are shown respectively in Table~\ref{tab:results-hier-phone-syllable} and Table~\ref{tab:results-hier-syllable-word}. Results are worse than when using either a hierarchical model with phone and word segments or a model with phone, syllable-word, and word segments.  This shows that preserving low-level segments such as phones and high-level segments such as words enables the model to better generalise. Also, according to the results presented in Table~\ref{tab:results-hier-phones-words}, it appears that the architecture of the network should be deeper (5 layers) when using both phone and word segments than when using other type of segments as the best models of Table~\ref{tab:results-hier-phone-syllable} and \ref{tab:results-hier-syllable-word} converge better with 4 layers. This suggests that using phone and word segments requires an additional amount of processing in order to be used effectively. Finally, the difference ($-0.3$pp) in the results obtained with a phone and word architecture, and a phone and syllable-word architecture show that even though syllables-word and words are quite close in length (see compression rates in section \ref{sec:seg-comp}), they are not equivalent in terms of semantic content, otherwise we would have observed identical results.
\begin{table*}[t]

% phones + syllables-word
	\resizebox{\textwidth}{!}{  
		\centering
		\begin{tabular}{|c|ccccc|cccc|ccc|cc|}
			\hline
			\textbf{Architecture} & \multicolumn{5}{c|}{\textbf{5 layers}}                                                                                                                                                    & \multicolumn{4}{c|}{\textbf{4 layers}}                                                                                           & \multicolumn{3}{c|}{\textbf{3 layers}}                                                         & \multicolumn{2}{c|}{\textbf{2 layers}}                       \\ \hline\hline
\diagbox{$1^{st}\text{\grupackager}$}
{$2^{nd}\text{\grupackager}$}     & \multicolumn{1}{c|}{\textbf{1}} & \multicolumn{1}{c|}{\textbf{2}} & \multicolumn{1}{c|}{\textbf{3}}                          & \multicolumn{1}{c|}{\textbf{4}} & \textbf{5}               & \multicolumn{1}{c|}{\textbf{1}} & \multicolumn{1}{c|}{\textbf{2}}   & \multicolumn{1}{c|}{\textbf{3}} & \textbf{4}               & \multicolumn{1}{c|}{\textbf{1}} & \multicolumn{1}{c|}{\textbf{2}}   & \textbf{3}               & \multicolumn{1}{c|}{\textbf{1}}   & \textbf{2}               \\ \hline
\textbf{1}                    & \multicolumn{1}{c|}{\cellcolor[HTML]{C0C0C0}} & \multicolumn{1}{c|}{6.6}                      & \multicolumn{1}{c|}{6.0}                      & \multicolumn{1}{c|}{6.3}                      & 4.3                      & \multicolumn{1}{c|}{\cellcolor[HTML]{C0C0C0}{ }} & \multicolumn{1}{c|}{{ 6.9}}                      & \multicolumn{1}{c|}{{ 6.5}}                               & { 4.6}                      & \multicolumn{1}{c|}{\cellcolor[HTML]{C0C0C0}{ }} & \multicolumn{1}{c|}{{ \textbf{6.6}}}             & { 4.9}                      & \multicolumn{1}{c|}{\cellcolor[HTML]{C0C0C0}{ }} & { \textbf{4.6}}             \\ \cline{1-1} \cline{3-6} \cline{8-10} \cline{12-13} \cline{15-15} 
			\textbf{2}                    & \cellcolor[HTML]{C0C0C0}                      & \multicolumn{1}{c|}{\cellcolor[HTML]{C0C0C0}} & \multicolumn{1}{c|}{{ \textbf{7.5}}}          & \multicolumn{1}{c|}{6.8}                      & 5.7                      & \cellcolor[HTML]{C0C0C0}{ }                      & \multicolumn{1}{c|}{\cellcolor[HTML]{C0C0C0}{ }} & \multicolumn{1}{c|}{{ \color[HTML]{FE0000} \textbf{7.9}}} & { 4.7}                      & \cellcolor[HTML]{C0C0C0}{ }                      & \multicolumn{1}{c|}{\cellcolor[HTML]{C0C0C0}{ }} & { 6.1}                      & \cellcolor[HTML]{C0C0C0}{ }                      & \cellcolor[HTML]{C0C0C0}{ } \\ \cline{1-1} \cline{4-6} \cline{9-10} \cline{13-13}
			\textbf{3}                    & \cellcolor[HTML]{C0C0C0}                      & \cellcolor[HTML]{C0C0C0}                      & \multicolumn{1}{c|}{\cellcolor[HTML]{C0C0C0}} & \multicolumn{1}{c|}{6.5}                      & 4.8                      & \cellcolor[HTML]{C0C0C0}{ }                      & \cellcolor[HTML]{C0C0C0}{  }                     & \multicolumn{1}{c|}{\cellcolor[HTML]{C0C0C0}{ }}          & { 4.7}                      & \cellcolor[HTML]{C0C0C0}{ }                      & \cellcolor[HTML]{C0C0C0}{ }                      & \cellcolor[HTML]{C0C0C0}{ } & \cellcolor[HTML]{C0C0C0}{ }                      & \cellcolor[HTML]{C0C0C0}{ } \\ \cline{1-1} \cline{5-6} \cline{10-10}
			\textbf{4}                    & \cellcolor[HTML]{C0C0C0}                      & \cellcolor[HTML]{C0C0C0}                      & \cellcolor[HTML]{C0C0C0}                      & \multicolumn{1}{c|}{\cellcolor[HTML]{C0C0C0}} & 4.6                      & \cellcolor[HTML]{C0C0C0}{ }                      & \cellcolor[HTML]{C0C0C0}{ }                      & \cellcolor[HTML]{C0C0C0}{ }                               & \cellcolor[HTML]{C0C0C0}{ } & \cellcolor[HTML]{C0C0C0}{ }                      & \cellcolor[HTML]{C0C0C0}{ }                      & \cellcolor[HTML]{C0C0C0}{ } & \cellcolor[HTML]{C0C0C0}{ }                      & \cellcolor[HTML]{C0C0C0}{ } \\ \cline{1-1} \cline{6-6}
			\textbf{5}                    & \cellcolor[HTML]{C0C0C0}                      & \cellcolor[HTML]{C0C0C0}                      & \cellcolor[HTML]{C0C0C0}                      & \cellcolor[HTML]{C0C0C0}                      & \cellcolor[HTML]{C0C0C0} & \cellcolor[HTML]{C0C0C0}{ }                      & \cellcolor[HTML]{C0C0C0}{ }                      & \cellcolor[HTML]{C0C0C0}{ }                               & \cellcolor[HTML]{C0C0C0}{ } & \cellcolor[HTML]{C0C0C0}{ }                      & \cellcolor[HTML]{C0C0C0}{ }                      & \cellcolor[HTML]{C0C0C0}{ } & \cellcolor[HTML]{C0C0C0}{ }                      & \cellcolor[HTML]{C0C0C0}{ } \\ \hline
			\textbf{Baseline}                & \multicolumn{5}{c|}{4.3}                                                                                                                                                                                                                                             & \multicolumn{4}{c|}{4.4}                                                                                                                                                                                                                                                               & \multicolumn{3}{c|}{3.4}                                                                                                                                                                               & \multicolumn{2}{c|}{3.5}                                                                                                                 \\ \hline
		\end{tabular}
	}
	\caption{R@1 obtained on the test set of the Flickr8k data set with a hierarchical architecture consisting of two \grupackager layers using phones and syllable-word (models were selected based on the maximum R@1 on the validation set). The same naming conventions of Table \ref{tab:results-hier-phones-words} are used for this table}
	\label{tab:results-hier-phone-syllable}
\end{table*}
\begin{table*}[!t]
% phones + syllables-word
	\resizebox{\textwidth}{!}{ 
		\centering
		\begin{tabular}{|c|ccccc|cccc|ccc|cc|}
			\hline
			\textbf{Architecture} & \multicolumn{5}{c|}{\textbf{5 layers}}                                                                                                                                                    & \multicolumn{4}{c|}{\textbf{4 layers}}                                                                                           & \multicolumn{3}{c|}{\textbf{3 layers}}                                                         & \multicolumn{2}{c|}{\textbf{2 layers}}                       \\ \hline\hline
\diagbox{$1^{st}\text{\grupackager}$}
{$2^{nd}\text{\grupackager}$}     & \multicolumn{1}{c|}{\textbf{1}} & \multicolumn{1}{c|}{\textbf{2}} & \multicolumn{1}{c|}{\textbf{3}}                          & \multicolumn{1}{c|}{\textbf{4}} & \textbf{5}               & \multicolumn{1}{c|}{\textbf{1}} & \multicolumn{1}{c|}{\textbf{2}}   & \multicolumn{1}{c|}{\textbf{3}} & \textbf{4}               & \multicolumn{1}{c|}{\textbf{1}} & \multicolumn{1}{c|}{\textbf{2}}   & \textbf{3}               & \multicolumn{1}{c|}{\textbf{1}}   & \textbf{2}               \\ \hline
\textbf{1}                    & \multicolumn{1}{c|}{\cellcolor[HTML]{C0C0C0}} & \multicolumn{1}{c|}{5.7}                      & \multicolumn{1}{c|}{5.5}                      & \multicolumn{1}{c|}{5.7}                      & 4.5                      & \multicolumn{1}{c|}{\cellcolor[HTML]{C0C0C0}{ }} & \multicolumn{1}{c|}{{ 5.7}}                      & \multicolumn{1}{c|}{{ 6.8}}                              & { 5.2}                      & \multicolumn{1}{c|}{\cellcolor[HTML]{C0C0C0}{ }} & \multicolumn{1}{c|}{{ 6.0}}                      & { 5.2}                      & \multicolumn{1}{c|}{\cellcolor[HTML]{C0C0C0}{ \textbf{}}} & { 5.3}                      \\ \cline{1-1} \cline{3-6} \cline{8-10} \cline{12-13} \cline{15-15} 
			\textbf{2}                    & \cellcolor[HTML]{C0C0C0}                      & \multicolumn{1}{c|}{\cellcolor[HTML]{C0C0C0}} & \multicolumn{1}{c|}{{ \textbf{7.3}}}          & \multicolumn{1}{c|}{7.1}                      & 6.1                      & \cellcolor[HTML]{C0C0C0}{ }                      & \multicolumn{1}{c|}{\cellcolor[HTML]{C0C0C0}{ }} & \multicolumn{1}{c|}{{\color[HTML]{FE0000} \textbf{7.6}}} & { 6.0}                      & \cellcolor[HTML]{C0C0C0}{ }                      & \multicolumn{1}{c|}{\cellcolor[HTML]{C0C0C0}{ }} & { \textbf{6.3}}             & \cellcolor[HTML]{C0C0C0}{ }                               & \cellcolor[HTML]{C0C0C0}{ } \\ \cline{1-1} \cline{4-6} \cline{9-10} \cline{13-13}
			\textbf{3}                    & \cellcolor[HTML]{C0C0C0}                      & \cellcolor[HTML]{C0C0C0}                      & \multicolumn{1}{c|}{\cellcolor[HTML]{C0C0C0}} & \multicolumn{1}{c|}{6.8}                      & 5.7                      & \cellcolor[HTML]{C0C0C0}{ }                      & \cellcolor[HTML]{C0C0C0}{ }                      & \multicolumn{1}{c|}{\cellcolor[HTML]{C0C0C0}{ }}         & { 6.0}                      & \cellcolor[HTML]{C0C0C0}{ }                      & \cellcolor[HTML]{C0C0C0}{ }                      & \cellcolor[HTML]{C0C0C0}{ } & \cellcolor[HTML]{C0C0C0}{ }                               & \cellcolor[HTML]{C0C0C0}{ } \\ \cline{1-1} \cline{5-6} \cline{10-10}
			\textbf{4}                    & \cellcolor[HTML]{C0C0C0}                      & \cellcolor[HTML]{C0C0C0}                      & \cellcolor[HTML]{C0C0C0}                      & \multicolumn{1}{c|}{\cellcolor[HTML]{C0C0C0}} & 5.5                      & \cellcolor[HTML]{C0C0C0}{ }                      & \cellcolor[HTML]{C0C0C0}{ }                      & \cellcolor[HTML]{C0C0C0}{ }                              & \cellcolor[HTML]{C0C0C0}{ } & \cellcolor[HTML]{C0C0C0}{ }                      & \cellcolor[HTML]{C0C0C0}{ }                      & \cellcolor[HTML]{C0C0C0}{ } & \cellcolor[HTML]{C0C0C0}{ }                               & \cellcolor[HTML]{C0C0C0}{ } \\ \cline{1-1} \cline{6-6}
			\textbf{5}                    & \cellcolor[HTML]{C0C0C0}                      & \cellcolor[HTML]{C0C0C0}                      & \cellcolor[HTML]{C0C0C0}                      & \cellcolor[HTML]{C0C0C0}                      & \cellcolor[HTML]{C0C0C0} & \cellcolor[HTML]{C0C0C0}{ }                      & \cellcolor[HTML]{C0C0C0}{ }                      & \cellcolor[HTML]{C0C0C0}{ }                              & \cellcolor[HTML]{C0C0C0}{ } & \cellcolor[HTML]{C0C0C0}{ }                      & \cellcolor[HTML]{C0C0C0}{ }                      & \cellcolor[HTML]{C0C0C0}{ } & \cellcolor[HTML]{C0C0C0}{ }                               & \cellcolor[HTML]{C0C0C0}{ } \\ \hline
			\textbf{Baseline}                & \multicolumn{5}{c|}{4.3}                                                                                                                                                                                                                                             & \multicolumn{4}{c|}{4.4}                                                                                                                                                                                                                                                               & \multicolumn{3}{c|}{3.4}                                                                                                                                                                               & \multicolumn{2}{c|}{3.5}                                                                                                                 \\ \hline
		\end{tabular}
	}
	\caption{R@1 obtained on the test set of the Flickr8k data set with a hierarchical architecture consisting of two \grupackager layers using syllable-word and word segments (models were selected based on the maximum R@1 on the validation set). The same naming conventions of Table \ref{tab:results-hier-phones-words} are used for this table.}
	\label{tab:results-hier-syllable-word}
\end{table*}
\end{document}